\documentclass[conference]{IEEEtran}
\IEEEoverridecommandlockouts
\usepackage{cite}
\usepackage{amsmath,amssymb,amsfonts}
\usepackage{algorithmic}
\usepackage{graphicx}
\usepackage{textcomp}
\usepackage{xcolor}
\usepackage{subfigure}
\usepackage{float}
\def\BibTeX{{\rm B\kern-.05em{\sc i\kern-.025em b}\kern-.08em
    T\kern-.1667em\lower.7ex\hbox{E}\kern-.125emX}}
\begin{document}

\title{Pneumothorax Segmentation: Deep Learning Image
Segmentation to predict Pneumothorax\\
}

\author{\IEEEauthorblockN{Karan Jakhar, Avneet Kaur, Dr Meenu Gupta}
\IEEEauthorblockA{\textit{Computer Science and Engineering Department
} \\
\textit{Chandigarh University}\\
Chandigarh, India \\
karanjakhar49@gmail.com,Avneet2900@gmail.com,meenu.e9406@cumail.in}
}

\maketitle

\begin{abstract}
Computer vision has shown promising results in the medical image processing. Pneumothorax is a deadly condition and if not diagnosed and treated at time then it causes death. It can be diagnosed from chest X-ray images. We need an expert and experienced radiologist to predict whether a person is suffering from pneumothorax or not by looking at the chest X-ray images. Everyone does not have access to such facility. Moreover, in some cases we need quick diagnoses. So we propose a image segmentation model to predict and give output a mask which will assist the doctor in taking this crucial decision. Deep Learning has proved their worth in many areas and outperformed many state-of-the-art models. We want to use the power of these deep learning model to solve this problem. We have used U-net\cite{b13} architecture with ResNet\cite{b17} as backbone and achieved promising results. U-net\cite{b13} performs very well in medical image processing and semantic segmentation. Our problem falls in the semantic segmentation category. 
\end{abstract}


\section{Introduction}
Pneumothorax is a life-threatening condition which is 
present in around 30-39\% of patients who are suffering from chest trauma \cite{b1,b2}. One way of diagnosing it with chest radiography. Pneumothorax is a condition in which air from lungs leaks to space between lungs and chest wall. This exert a force on the lungs and stop them from expanding.Due to this the person can not breathe and if not treated then it causes death. However, it is a very challenge task and often fail to identify a pneumothorax and even reaching false-negative rate to 50\%\cite{b3,b4}. If we fail to treat an enlarging pneumothorax in time then it may result in patient death. Chest X-ray is a way to diagnose if a patient is have pneumothorax or not. But to do so we need well qualified and experienced radiologists. Everyone can not have access to such facility. So we want to use technology here. Today, technology have power to solve any problem. We just need to use it in that way. Advancement in Deep Learning in recent year make it is to work with image data. Chest X-ray images will be processed by our model and it will predict whether a person is suffering from pneumothorax or not. We also want to perform image segmentation on it and give a mask as output so that doctor could know which area is problematic. This also gives a sense to doctor why the model is predicting so. 
In recent years, deep convolution networks have
outperformed other the state of the art model in many visual recognition
tasks, e.g. \cite{b7,b8}. Though convolutional networks were already
there for a long time \cite{b9} but  their success was limited as we didn't have
large amount of data and computation power to handle large networks. In 2012 Krizhevsky et al.\cite{b10} make a breakthrough by supervised training of a large network having 8
layers and millions of parameters on the ImageNet dataset which have 1 million training images. After that many deeper networks have been introduced and a lot of work has been done on it \cite{b11}. Having a model which can predict the weather a person is have Pneumothorax problem or not is very helpful for doctors to start the treatment at the earliest. Deep learning is a very powerful tool and we should use it’s power to solve the problems which can save some lives. We prepared a model which can be easily deployed and used by the doctors to get real time results. It will not only predict weather a person is have pneumothorax but also indicated the problem area. To predict if a patient is suffering from Pneumothorax just by read the X-rays required a lot of expertise in the field. We need experience and well qualified radiologists to do this. But everyone doesn’t have access to a radiologist and it is expensive also\cite{b12}. This process of prediction by a human being is very time consuming.

\begin{figure}[h]
\includegraphics[width=\linewidth]{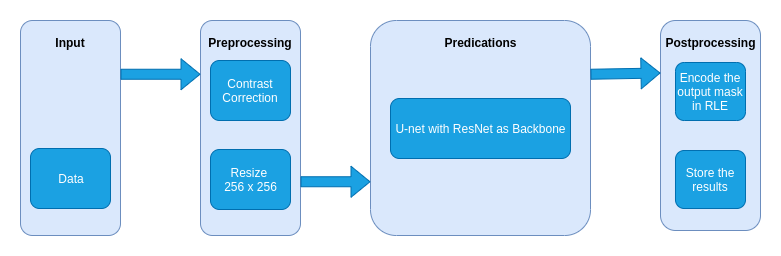}
\caption{Proposed Framework}
\label{fig:framework}
\end{figure}

Using technology to solve medical problems is great way to utilize its power. Today, we have such strong models and hardware to solve many problems which seems impossible without them. We should develop such systems which can assist doctors to take crucial decisions. We propose a system which will help the doctors to take decision whether a patient has pneumothorax or not. Our model will give a mask showing the problematic area due to which our model is predicting that patient has pneumothorax. After having a look at the mask the doctor can easily take decision whether the patient is suffering from pneumothorax and what treatment is required.

\section{Related Work}
There is a lot of work done in the field of Image Segmentation
and Pneumothorax independently. Image Segmentation is used
before in medical imaging but it is not yet used for
Pneumothorax. We have U-net\cite{b13} which is used in ISBI
challenge and won the competition. The task was to segment  neuronal structures present in  electron microscopic stacks. Segmentation of a 256 X 256 or greater can be done in microseconds on a recent GPU. Our problem falls in category of semantic segmentation\cite{b24}. Before deep learning methods there were many techniques used to segment an object in an image for example region based, edge detection etc. RCNN\cite{b25} in 2015 used deep learning architecture to successfully achieve state-of-the-art results at that time. Its main focus was on object detection and drawing bounding boxes around the object. Later on it was improved further with introduction of Fast-RCNN\cite{b26} after that we had Faster-RCNN\cite{b27}. In 2018 FAIR(Facebook AI Research) introduced Mask-RCNN\cite{b28} which has amazing results on COCO dataset\cite{b29}. Its main focus was on Instance Segmentation. There are many image segmentation
techniques are available in the literature. U-net has been used a lot in medical imaging as it shows promising results. U-net\cite{b13} uses concept of skip connection as described in ResNet model architecture\cite{b17}. With skip connection we avoid problem of vanishing gradient as we go deeper and also preserve the more information when upsampling. U-net works on the concept of downsampling and upsampling like encoder-decoder models. Mask R-CNN is simple and addition to Faster R-CNN which generate two outputs corresponding to each candidate object, a class label and a bounding-box offset and in case of Mask R-CNN we add a third branch which generates the object mask. I am combining best work from both fields and produce useful results which can be used by doctors to diagnose Pneumothorax as early as possible and keeping false-negative low. U-net is more related to the generative models with little difference that is instead of learning to generate the same image, it learns to generate the mask. A lot of work and experimentation is done on generative models. We have guidelines to improve generative models. Pneumothorax is a well known disease and it is very tough in some situations for a doctor to come to a conclusion that the person is suffering from Pneumothorax. That’s why having a deep learning model assisting the doctor in taking decision is very helpful. As this result will be a mask at the problematic area so the doctor can consider his decision easily. 

\section{Material and Methods}

\subsection{Proposed Framework}
The output of  U-net model with pretrained weights of the ResNet is fast and accurate. This helps doctors to start the treatment at the earliest. U-net is consist of an encoder and decoder. Encoder narrow down the information of the image to a latent space. Then decoder taking input latent space regenerates the image. Here the model will be generating the mask. The mask tells us were is Pneumothorax if it exists. Image preprocessing is an important part here. As real world images are not of the same size and more sometime are corrupted. In preprocessing we have to resize the image in 256 X 256. Then the image is put into correct color range. The image will be normalized. For training we filtered out the corrupted images and done random crop to remove unwanted boundary. As X-ray images usually have a lot of boundary noise which is not required for our model. The output mask is a black and white image. In which white area indicated the position of the problem. We then map this mask image to out original image to get exact position on the X-ray image. The encoder part downscale the images and learn a latent space with help of which we generate the masks.  This is kind of mapping an image to a mask and training our model to learn to generalist the mapping. When the model successfully learns to generalize it, we can feed any X-ray image and get the mask for it. The mask will indicate the position of the part which is potentially causing Pneumothorax. If the mask is blank means there is no problem and patient is safe. Prediction should be very fast because if it take too long then it can  not help the doctor in needed time. U-net is pretty small network and very effective also. We are using ResNet as encoder and we are going to use the weights from pretrained network. Using weights from pretrained network helps to boost the result and saves us time. Skip connections of U-net help to retain useful features learned in encoder layer while upsampling by decoder. ResNet concept of skip connection helps fight vanishing gradient problem. That’s why decided to try the ResNet pretrained model as backbone. It improves the result.

\begin{figure}
  \includegraphics[width=\linewidth]{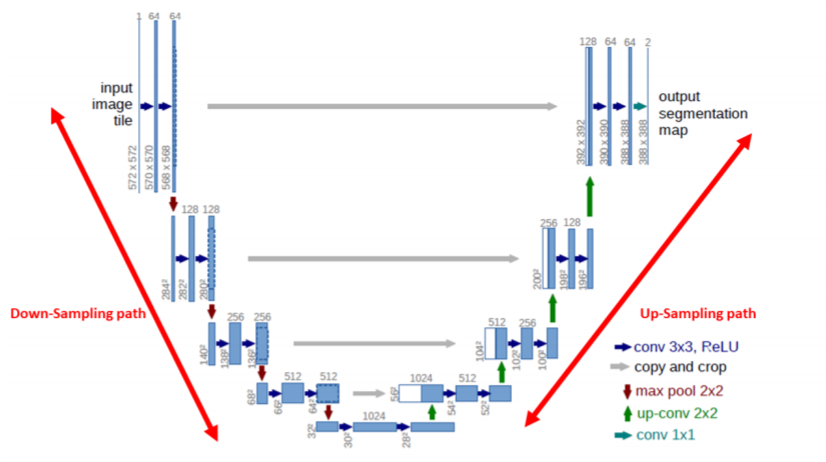}
  \caption{Classical U-net Architecture}
  \label{fig:classical_unet}
\end{figure}

\begin{figure}
  \includegraphics[width=\linewidth]{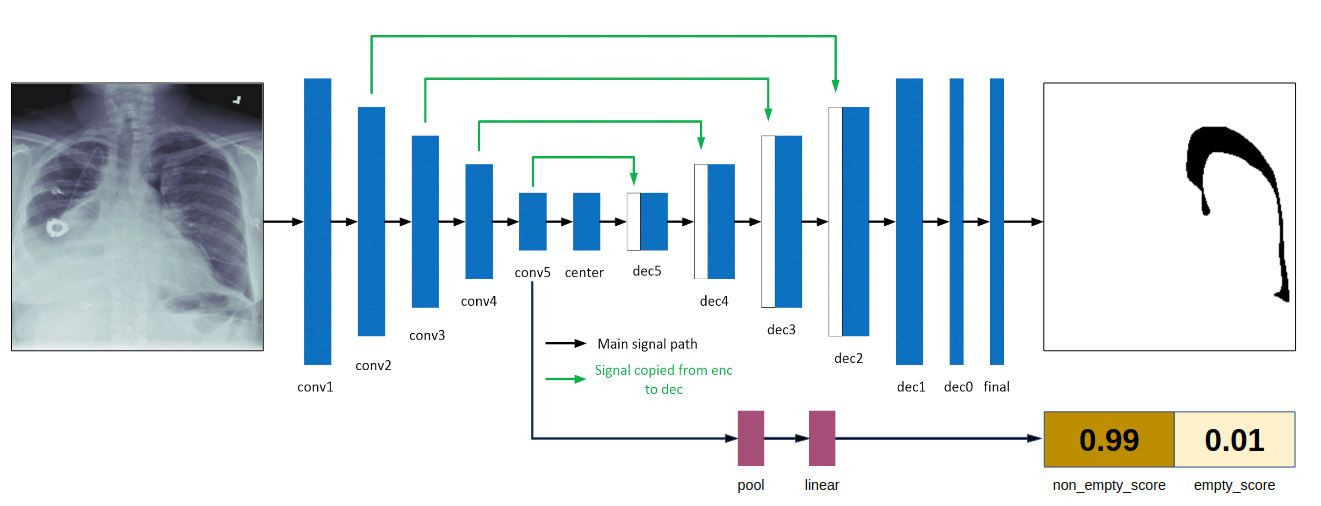}
  \caption{U-net Architecture Used}
  \label{fig:model}
\end{figure}

\subsection{Segmentation Techniques }
 There has been a lot of shift in approaches and techniques used to perform segmentation. Before Deep Learning was a thing, various methods were adapted which were not efficient and were not giving the required results. With Deep Learning entering into the picture various architectures have been developed. Unet is one such architecture concept which works well in medical image segmentation. Not every architecture works best for every problem. There have been many shifts in state-of-the-art models in image segmentation after the arrival of deep learning. Image Segmentation falls into object detection problem. Earlier researchers found solutions which will find the object in the image, predict its class and draw a bounding box around it. In Image Segmentation we predict a masks for an object in the image. Below we will discuss a few of the approaches used for segmentation. There are different type of segmentation such as semantic segmentation and instance segmentation. Our problem falls into semantic segmentation. 
\paragraph{Semantic Segmentation}
  
In this we classify each and every pixel of the image into a class. Object belonging to same class will be given same colour. It does not classify objects belonging to same class differently. 
Instance Segmentation   
It is different then Semantic Segmentation in context that if classify each object in the image differently whether it belongs to same class or not. Every object gets a label to it. It is also we tough to achieve and require a lot of different techniques to get desired results.

\paragraph{Region based segmentation}
  
In this type of segmentation we divide regions with its pixel value. This works well when we have a sharp contrast between object and background. We can set a threshold value and pixel values falling below and above it can to categorized accordingly. This technique is called threshold segmentation\cite{b30}.

\paragraph{Edge Detection}
    
We use filters and convolutions to identify the edge or boundary between two region. When we move from one region to another in an image they have a edge that separates them. If we can find that edge then it is very easy to separate them. We use filters for that. We have different types of filters which can detect a horizontal, vertical or both edges.

\paragraph{Clustering}

In this we divided the polulation into different groups. We can divided pixels in different groups and like this we can segment them. For example we can use K-Means clustter algorithm and can easily form different clusters of similar data points or pixels. It is not suitable for large amount of data.

\paragraph{FCN(fully convolutional network) }
  
This type of model uses convolutional and pooling layers to map an image to masks. These masks are used to produce the segmented image. The concept is simple. First image is downsampled using convolutional and pooling layers then upsampled in to the same size but generating the masks\cite{b14}. 

\paragraph{Mask R-CNN}

This is state-of-the-art model for instance segmentation. It extends Faster R-CNN[16]. Faster R-CNN\cite{b16} gives bounding box and label for the class but Mask R-CNN\cite{b15} has added another branch which will give mask of the object. The concept is simple first we extract feature out of image then regions are proposed. Then we find region of interest. Going according to this process the model generates label, bounding box, and mask for the object.

\section{Experimental Investigation}

\subsection{Dataset}
The dataset download from Kaggle\cite{b18} which has training image and their masks. The masks are encoded in RLE(Run Lenght Encoding) and we have used the same algorithm to decode the mask image. We kept some of the training images for testing and validating the model after training. Dataset has good amount of the images. In the dataset we have 12047 images and same number of masks. Images are in shape 256 x 256. If an image does not have Pneumothorax then it’s mask is all black. The all black mask image teaches the model which X-ray images does not have Pneumothorax. We will be keeping 20\% of the images for testing and validating the model. 
\begin{figure}[h]
\includegraphics[width=\linewidth]{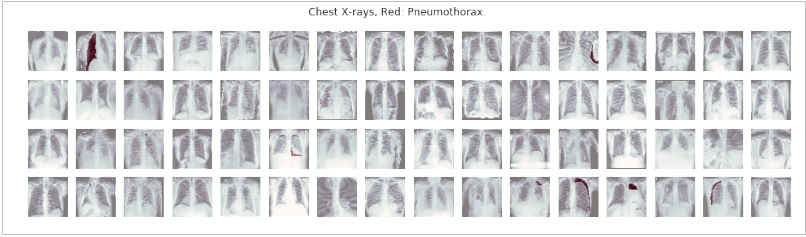}
\caption{Chest X-ray Images}
\label{fig:dataset}
\end{figure}
\subsection{Experimental Setting}
Python programming language is simple, fast and efficient way to implement deep learning models. We used Keras\cite{b22} library in Python which is high level API. Keras is fast and at the backend it uses Tensorflow\cite{b21}. Due to Keras we need not to write a lot of repetitive code. Image processing is also a great part of the experiment. We used OpenCV\cite{b23} to process the image. The major operation we performed is contrast correction. The code becomes concise and easy to interpret. The model is implemented in parts. First encoder, then decoder is implemented. We used pretrained weights of ResNet model. Therefore, ResNet is used as encoder model. Without pretrained weights the result is not good. Deep learning models are data hungry and need a lot of computation power to train. As we are training deep learning model we need a lot of computation power. We need good GPU. It is very time consuming to train the model on a CPU. We used Adam optimizer with learning rate 0.0001, binary cross-entropy loss and early stop validation. To get a clear output mask we used a threshold techniques to remove the noise and save only highly supported pixel values. Without setting threshold and filtering pixels we get little noise in some mask and output masks are not that sharp. Careful post-processing is also required. We used OpenCV\cite{b23} to perform this process. 
\section{Results and Discussion}
\subsection{Performance Evaluation}
\begin{figure}[ht]
\centering 
  \subfigure[Dice Coefficient]{%
    \includegraphics[width=0.4\linewidth]{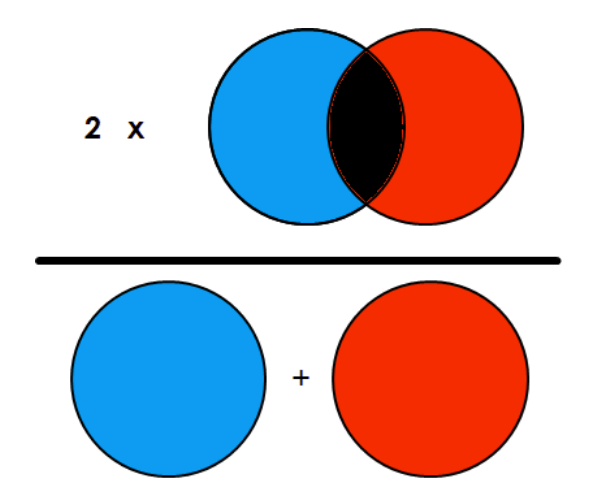} \label{fig:1} 
  } 
  \quad 
  \subfigure[IoU]{%
    \includegraphics[width=0.4\linewidth]{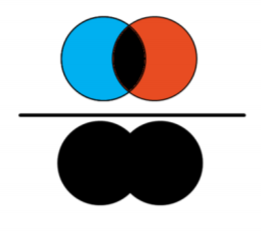} \label{fig:2} 
  } 
  \caption{Metrics Illustrations} 
\end{figure}

Choosing right metric is very important to evaluate a model. We evaluated our model on Dice Coefficient\cite{b19}, IOU\cite{b20} which are considered best metrics to evaluate a segmentation result. There are various metrics to evaluate segmentation. But we are using dice coefficient as described in \cite{b19}. Checking result by visualizing is good but when we are train we need some metrics to monitor the model. We use dice and IoU because are proven best when we have to check similarity between two images. And here we want to check how much our output mask is similar to real mask. The formulae of these score metrics is easy to understand and are shown in figures \ref{fig:1} \ref{fig:2}.

\begin{equation}
D =  \frac{2|X \cap Y|}{|X| + |Y|} \label{eq:1}
\end{equation}

\begin{equation}
IoU =  \frac{|X \cap Y|}{|X \cup Y|} \label{eq:2}
\end{equation}

\subsection{Experimental Results}

\begin{figure}[H]
\centering 
  \subfigure[Input image]{%
    \includegraphics[width=0.2\linewidth]{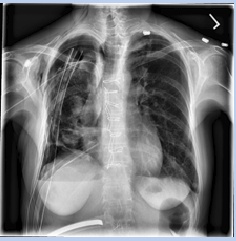} \label{fig:3} 
  } 
  \quad 
  \subfigure[Image with predicted mask]{%
    \includegraphics[width=0.2\linewidth]{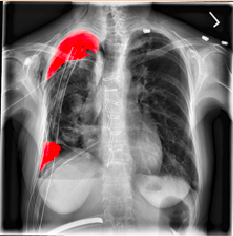} \label{fig:4} 
  } 
  \quad 
  \subfigure[Real Annotation]{%
    \includegraphics[width=0.2\linewidth]{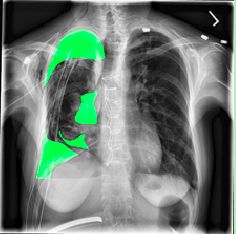} \label{fig:5} 
  } 
  \caption{Patient 1} 
\end{figure}

\begin{figure}[H]
\centering 
  \subfigure[Input image]{%
    \includegraphics[width=0.2\linewidth]{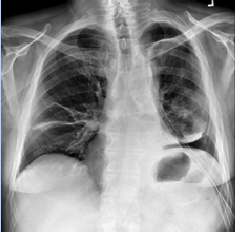} \label{fig:6} 
  } 
  \quad 
  \subfigure[Image with predicted mask]{%
    \includegraphics[width=0.2\linewidth]{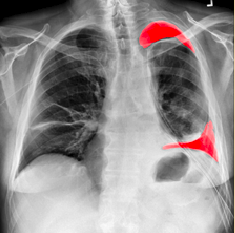} \label{fig:7} 
  } 
  \quad 
  \subfigure[Real Annotation]{%
    \includegraphics[width=0.2\linewidth]{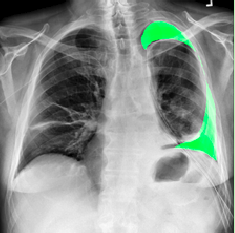} \label{fig:8} 
  } 
  \caption{Patient 2} 
\end{figure}

\begin{figure}[H]
\centering 
  \subfigure[Input image]{%
    \includegraphics[width=0.2\linewidth]{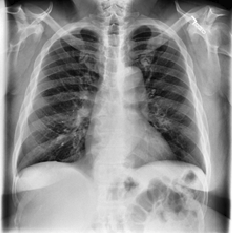} \label{fig:9} 
  } 
  \quad 
  \subfigure[Image with predicted mask]{%
    \includegraphics[width=0.2\linewidth]{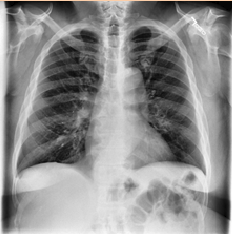} \label{fig:10} 
  } 
  \quad 
  \subfigure[Real Annotation]{%
    \includegraphics[width=0.2\linewidth]{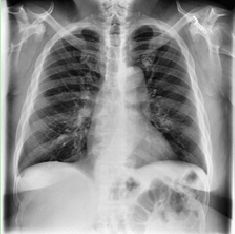} \label{fig:11} 
  } 
  \caption{Patient 3} 
\end{figure}

Our model is showing promising results as we can see the masks predicted by the model in figures \ref{fig:3} \ref{fig:4} \ref{fig:5} match closly to the real annotation. Our Dice score and IoU score are 84.3 and 82.6 respectively. Both of these metrics are used to indicate the similarity between images. And in figures \ref{fig:6} \ref{fig:7} \ref{fig:8} \ref{fig:9} \ref{fig:10} \ref{fig:11} we can see the results are very good and insightful. Both Dice score and IoU indicates that model is strong. Dice score and IoU both are positively correlated. It means if one say a model is strong then other will also support the decision. Using ResNet as backbone in U-net architecture helps converse the model fast and improves the results. The implementation is also easy in Keras. There are other metrics also but these metrics are considered best to test similarity between two images.

\begin{table}[H]
\caption{Scores}
\begin{center}
\begin{tabular}{|c|c|}
\hline
Dice Coefficient & IoU \\
\hline
84.3 & 82.6 \\
\hline
\end{tabular}
\label{tab1}
\end{center}

\end{table}

\begin{figure}[H]
\includegraphics[width=\linewidth]{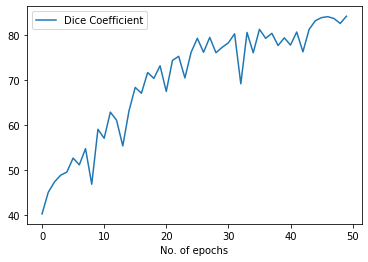}
\caption{Dice Coefficient Valiadation score}
\label{fig:dice_val}
\end{figure}

\begin{figure}[H]
\includegraphics[width=\linewidth]{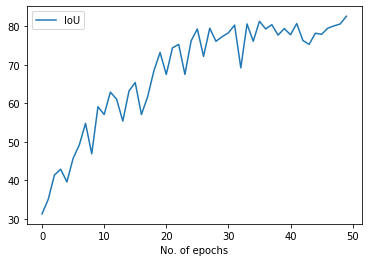}
\caption{IoU Valiadation score}
\label{fig:dice_val}
\end{figure}

\section*{Conclusion and Scope}

Deep learning is a powerfull tool. If it can be used effectively then it could help us to tackle a lot of problem. Using segmentation to predict Pneumothorax provides a great help to doctors to deal with it. Many people die due to Pneumothorax because it took a lot of time to diagnose the problem. With help of this great technology we can diagnose faster and with great accuracy. Our proposed model is fast and accurate. It can easily be deployed and used. Using U-net with pretrained ResNet weights is a effective way to get a great model in less time. With advancement in hardware we can make this prediction in seconds. This accurate and fast result will help doctors a lot to take decision and start their treatment. The model performs very well and training time is also decreased due to ResNet pretrained weights. Using ResNet as backbone helps model to converse fast and it also deals with various problems also like vanishing gradient. Computer vision has many applications in medical field this is just one of them. There are many to explore further.

\end{document}